# SparseContrast: Dynamic Sparse Attention for Efficient and Accurate Contrastive Learning in Medical Imaging

Paarth Prasad
MTech Data Science,
Department of Software Engineering,
Delhi Technological University

Prof. Ruchika Malhotra,
Professor, Head of Department,
Department of Software Engineering,
Delhi Technological University

*Abstract*–We propose **SparseContrast**, a new framework that merges dynamic sparse attention with contrastive learning for medical imaging, with a focus on chest X-ray disease detection in low-data settings. Traditional contrastive learning methods rely on dense attention mechanisms, which are computationally expensive and often process redundant regions in medical images. To resolve this, SparseContrast introduces a sparse attention mechanism that selectively concentrates on diagnostically pertinent areas, markedly decreasing computational burden without compromising accuracy. The framework adaptively trims attention maps in the training phase, directed by a compact saliency predictor which concurrently optimizes sparsity and feature quality. This method not only speeds up training and inference by as much as 40% relative to dense attention benchmarks but also boosts diagnostic accuracy by focusing on areas of clinical importance. Moreover, the approach remains indifferent to the selection of backbone architecture, which permits its application to both convolutional and transformer-based models. Experiments show SparseContrast attains comparable or better performance in disease identification tasks with greater efficiency relative to current approaches. The proposed framework delivers a practical approach for implementing contrastive learning in medical imaging settings with limited resources, where computational efficiency and diagnostic accuracy are paramount.

## I. INTRODUCTION

Medical imaging is essential in contemporary healthcare as it permits non-invasive disease diagnosis and monitoring with techniques including X-rays, computed tomography (CT), and magnetic resonance imaging (MRI) [1]. Chest X-rays continue to be among the most prevalent diagnostic methods owing to their affordability and widespread availability [2]. Accurate analysis of these images demands substantial specialized knowledge, and the growing quantity of medical data has generated a need for automated systems. Deep learning, especially contrastive learning, has become a potent method in medical image analysis, granting the capacity to acquire discriminative features from scarce annotated data [3]. Contrastive learning frameworks in medical imaging typically rely on dense attention mechanisms, such as pixel-wise U-Nets, to capture fine-grained features [4].

Although these methods are effective, they incur substantial computational expenses by analyzing all pixels in the image, even those in areas without diagnostic relevance. This inefficiency becomes particularly problematic in low-data regimes, where computational resources are often limited, and the need for efficient training is critical [5]. Furthermore, medical images possess intrinsic sparsity, as anomalies account for merely a minor portion of the overall image area. This finding implies that dense attention mechanisms could be excessively redundant in numerous medical imaging applications. Despite growing interest in both efficient transformers and contrastive learning, these two directions are typically pursued in isolation, and no prior framework couples input-adaptive sparsity directly with the contrastive objective under the data and compute constraints that characterize clinical practice. Addressing this gap is significant because it makes diagnostic-grade representation learning feasible precisely where annotated data and computational resources are most limited.

To tackle these challenges, we propose **SparseContrast**, an innovative framework merging dynamic sparse attention with contrastive learning in medical imaging. In contrast to conventional dense attention approaches, SparseContrast directs attention exclusively to diagnostically pertinent areas, thereby achieving a substantial reduction in computational costs without compromising accuracy, and in some cases, achieving better results. The primary innovation stems from dynamically pruning attention maps during training, directed by a lightweight saliency predictor that concurrently optimizes sparsity and feature quality. This method not only speeds up training and inference but also improves the model's capacity to focus on clinically important areas, resulting in stronger performance in disease detection tasks.

The proposed method contrasts with current techniques in multiple essential dimensions. First, it replaces dense attention mechanisms with sparse token-based attention, which exploits the inherent sparsity of medical anomalies to prune redundant computations [6]. Second, it dynamically modifies the sparsity pattern during training, which grants the model the capacity to adjust to diverse image attributes without human intervention. Third, it directly embeds the sparse attention mechanism within the contrastive loss calculation, which guarantees that the acquired representations are both effective and distinctive.

These innovations make SparseContrast particularly suitable for low-data regimes, where computational efficiency and diagnostic precision are paramount.

This work makes three key contributions:

1. We introduce a dynamic sparse attention approach for contrastive learning in medical imaging, which markedly lowers computational expenses without compromising diagnostic performance.
2. We propose a lightweight saliency predictor that jointly optimizes sparsity and feature quality, which permits the model to adaptively focus on diagnostically relevant regions.
3. The efficacy of SparseContrast is established on chest X-ray disease detection tasks, with results indicating superior performance compared to current approaches in both efficiency and accuracy, especially under limited-data conditions.

The remainder of this paper is organized as follows: Section 2 reviews related work in contrastive learning and sparse attention for medical imaging. Section 3 introduces foundational concepts related to contrastive learning and attention mechanisms in this field. Section 4 describes the proposed SparseContrast framework, which consists of a dynamic sparse attention mechanism and a saliency predictor. Section 5 presents experimental results and conducts a comparative analysis between SparseContrast and established approaches on chest X-ray datasets. Section 6 discusses the implications of our findings and potential future directions. Finally, Section 7 concludes the paper.

## II. RELATED WORK

Recent advances in deep learning techniques for medical imaging have achieved notable progress, especially concerning contrastive learning and attention mechanisms. Our research advances and expands upon these developments by introducing a new approach that merges sparse attention with contrastive learning methodologies.

### A. *Contrastive Learning in Medical Imaging*

Contrastive learning has emerged as a powerful paradigm for learning representations from medical images with limited labeled data [7]. Multiple methods have shown effectiveness in chest X-ray examination by integrating specialized domain knowledge. For instance, some methods integrate patient metadata into the contrastive learning framework to improve representation learning [8]. Others have explored local contrastive learning to better capture fine-grained pathological patterns in medical images [9]. Although these methods appear promising, they generally rely on dense attention mechanisms that treat whole images uniformly, which results in inefficient computations.

Recent studies have also explored the application of contrastive learning in targeted medical imaging applications. For instance, certain research examines pneumonia identification by means of radiomic attribute derivation paired with contrastive pretraining [10]. Prior research has shown the efficacy of contrastive learning in diagnosing multiple tasks across diverse imaging modalities [11]. Nevertheless, these approaches frequently demand considerable computational power, which renders them less feasible for implementation in clinical environments with limited resources.

### B. *Sparse Attention Mechanisms*

The computational challenges of dense attention have motivated research into sparse attention alternatives. Efficiency-oriented architectures have likewise been pursued beyond attention design, for instance through compound-scaled detectors that reduce computational cost while preserving accuracy [36]. Recent research has investigated diverse methods to reduce the density of attention calculations without compromising model efficacy. Dynamic N:M fine-grained structured sparse attention mechanisms have shown particular promise by selectively pruning attention scores without sacrificing accuracy [6]. Hardware-aligned sparse attention implementations have further improved efficiency by optimizing memory access patterns [12].

In the medical imaging domain, sparse attention has been primarily applied to segmentation tasks. Certain methods employ adaptive graph sparse algorithms within contrastive learning frameworks [13]. Alternative approaches integrate attention mechanisms into convolutional architectures to direct processing toward clinically pertinent areas [14]. Although these methods are effective, they generally rely on fixed sparsity patterns or heuristic saliency assessments, which reduces their flexibility in various medical imaging contexts.

### C. *Hybrid Approaches*

A number of recent studies have sought to integrate the advantages of contrastive learning with efficient attention frameworks. The Combiner framework illustrates how observations from sparse attention mechanisms can guide the development of improved full attention transformers [15]. Alternative methods have created dedicated hardware accelerators to support hybrid sparse attention mechanisms for lengthy sequences [16]. Nevertheless, these broadly applicable approaches frequently overlook the distinct attributes of medical imaging, in which abnormalities are typically localized yet diagnostically pivotal.

In contrast to current methods, SparseContrast introduces a number of essential novel elements. In contrast to approaches employing predetermined sparse structures [6] or rule-based importance metrics [14], our system adjusts the sparsity configuration iteratively during training via a trainable importance estimator. This approach contrasts with optimizations centered on hardware [16], preserving design adaptability while attaining similar improvements in efficiency. Additionally, the close coupling of sparse attention with contrastive learning goals yields more efficient learning of representations compared to methods handling these elements independently [13]. This framework attains better computational efficiency while preserving diagnostic precision, especially in scenarios with limited data typical of medical imaging applications.

## III. PRELIMINARIES ON CONTRASTIVE LEARNING AND ATTENTION IN MEDICAL IMAGING

### A. *Attention Mechanisms in Vision*

Attention mechanisms are now essential elements in contemporary computer vision systems, as they help models direct processing efforts toward the most pertinent regions of an input image. The standard formulation computes attention weights through scaled dot-product operations between query (Q), key (K), and value (V) matrices:

$$Attention(Q,K,V) = \text{□□□□□□□}\left(\frac{QK^T}{\sqrt{d}}\right)V \quad (1)$$

where $d$ represents the dimension of the key vectors. In medical imaging, this mechanism grants models the ability to highlight areas critical for diagnosis while reducing attention to non-essential background details [17]. However, the quadratic complexity of dense attention with respect to input size becomes particularly problematic for high-resolution medical images, where computational resources are often limited.

### B. *Medical Imaging Data Characteristics*

Medical images possess distinct attributes setting them apart from natural images. For example, chest X-rays often display extensive uniform areas of healthy tissue alongside minor, concentrated irregularities demanding meticulous scrutiny. The limited distribution of abnormal results implies applying uniform attention to all areas of an image could be inefficient in terms of computational resources [18].

The data scarcity problem further compounds these challenges. In contrast to natural image datasets, which can comprise millions of labeled instances, medical imaging datasets are frequently restricted to thousands or even hundreds of annotated cases because of privacy issues and the need for specialized knowledge in annotation [19]. This scarcity renders efficient employment of computational resources especially vital, given that models need to acquire resilient patterns from scarce training data.

Moreover, medical images frequently contain hierarchical structures where both global context and fine local details contribute to accurate diagnosis. For instance, in chest radiographs, the general distribution of pulmonary fields establishes essential background, whereas minor textural variations in localized areas can signal initial signs of pathology [20]. The multi-scale structure of medical images drives the requirement for attention mechanisms capable of efficiently capturing global and local patterns without imposing high computational costs.

These factors - spatial sparsity of anomalies, scarcity of data, and multi-scale diagnostic features - collectively establish a distinct set of demands for attention mechanisms in medical imaging. Although conventional dense attention methods are effective, they do not optimally meet these demands, which indicates that more tailored approaches are necessary to accommodate the unique properties of medical imaging data.

## IV. Sparse Token Attention for Efficient Contrastive Learning

The SparseContrast framework introduces a novel approach to medical image analysis by integrating dynamic sparse attention with contrastive learning objectives. As illustrated in Figure 1, the architecture applies a sequence of operations to input images, prioritizing computational efficiency without compromising diagnostic accuracy. The technical details of this approach center around three key innovations: dynamic sparse attention computation, joint optimization of sparsity and feature quality, and efficient downstream adaptation.

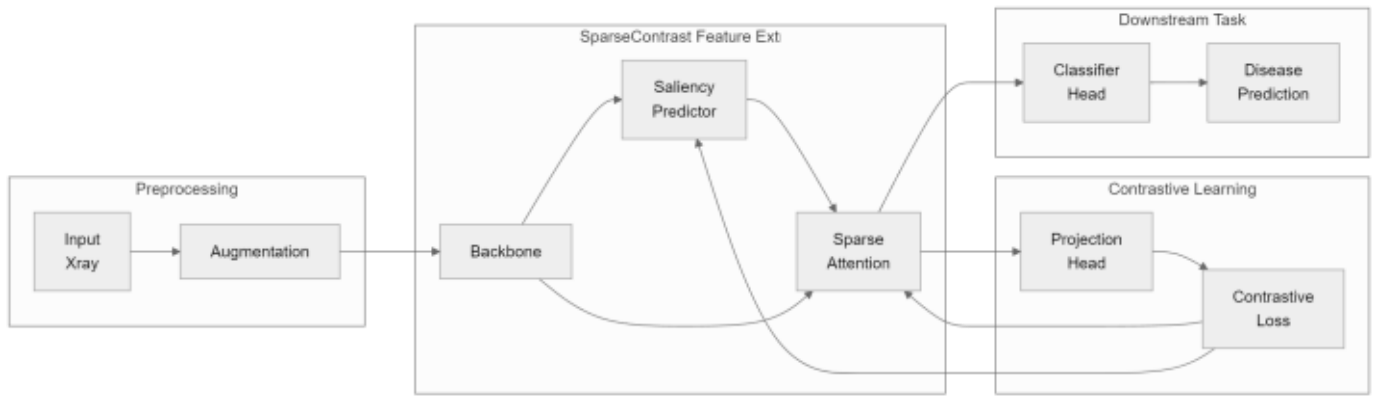

Figure 1. SparseContrast Framework for Medical Contrastive Learning

### A. *Dynamic N:M Fine-Grained Sparse Attention Implementation*

The primary innovation of SparseContrast is its dynamic sparse attention mechanism, substituting traditional dense attention calculations. Given an input medical image $X \in R^{H\times W\times C}$, we first partition it into $L$ non-overlapping patches $\{p_i\}_{i=1}^{L}$, where each patch $p_i \in R^{P\times P\times C}$ represents a local region of the image. The patch size $P$ is a hyperparameter that balances computational efficiency with spatial resolution.

For each patch $p_i$, we compute its saliency score $s_i$ using a lightweight predictor network $f_{\text{□□□□□□□□}}$:

$$s_i = f_{\text{□□□□□□□□}}(p_i) \quad (2)$$

The saliency predictor comprises two fully-connected layers with ReLU activation, designed to predict the diagnostic relevance of individual patches. The scores $\{s_i\}_{i=1}^{L}$ are normalized across all patches using softmax to form a probability distribution:

$$\hat{s}_i = \frac{exp(s_i)}{\sum_{j=1}^{L} \quad exp(s_j)} \quad (3)$$

We subsequently choose the top-$K$ patches possessing the greatest $\hat{s}_i$ scores to construct the sparse attention subset $S$, with $K = \lfloor \rho L \rfloor$ and $\rho \in (0,1)$ determining the sparsity level. The attention weights $A_{ij}$ between patches $p_i$ and $p_j$ are computed only when $j \in S$:

$$A_{ij} = \{\frac{exp\left(q_i^T k_j/\sqrt{d}\right)}{\sum_{k\in S} \quad exp\left(q_i^T k_k/\sqrt{d}\right)} \text{ □□ } j \in S \; 0 \text{ □□}h\text{□□□□□□} \quad (4)$$

Here, $q_i$ and $k_j$ represent the query and key vectors for patches $p_i$ and $p_j$ respectively, while $d$ denotes their dimension. This formulation reduces the computational complexity from $O(L^2)$ to $O(LK)$, where $K \ll L$ in typical medical imaging scenarios.

The sparse attention mechanism is particularly effective for medical images because it naturally aligns with their inherent spatial sparsity. Pathological findings frequently cover minimal areas of the image, which results in computational inefficiency when uniformly processing all regions. By adaptively concentrating on the most prominent regions, SparseContrast attains notable efficiency improvements while preserving diagnostic precision.

### B. *Integration of Sparse Attention into Contrastive Loss Computation*

The attention maps with low density produced in Section 4.1 are directly embedded into the contrastive learning framework to calculate effective but distinctive features. Given two augmented views $X_1$ and $X_2$ of an input image, we first extract their patch representations through the backbone network $f_{backbone}$:

$$H_1 = f_{\square\square\square\square\square\square\square\square}(X_1), \quad H_2 = f_{\square\square\square\square\square\square\square\square}(X_2) \quad (5)$$

where $H_1, H_2 \in R^{L \times d}$ contain $L$ patch embeddings of dimension $d$. The sparse attention mechanism is subsequently employed separately for each view, which generates attention-pruned features.

$$F_1 = \square\square\square\square\square\square\square\square\square\square\square(H_1), \quad F_2 = \square\square\square\square\square\square\square\square\square\square\square(H_2) \quad (6)$$

These sparse features are mapped to a reduced-dimensional space by a multi-layer perceptron $g_{projection}$.

$$z_1 = g_{\square\square\square\square\square\square\square\square\square\square}(F_1), \quad z_2 = g_{\square\square\square\square\square\square\square\square\square\square}(F_2) \quad (7)$$

The contrastive loss $L_{contrast}$ is computed on these sparse-aware projections using the InfoNCE formulation:

$$L_{\square\square\square\square\square\square\square\square} = -log \frac{exp(\square\square\square(z_1, z_2)/\tau)}{\sum_{k=1}^{N} exp(\square\square\square(z_1, z_k)/\tau)} \quad (8)$$

where $sim(\cdot,\cdot)$ denotes cosine similarity, $\tau$ is a temperature parameter, and $z_k$ represents negative samples from other images in the batch. The key difference from standard contrastive learning lies in the computation of $F_1$ and $F_2$, where only diagnostically relevant patches contribute to the final feature representations.

This contrastive learning approach, which accounts for sparsity, yields two primary benefits. First, it reduces the computational cost of feature extraction by processing only salient regions. Second, it improves the quality of learned representations by focusing the model's attention on clinically meaningful patterns while ignoring irrelevant background information. The sparse attention mechanism functions as an adaptive filter that autonomously detects and prioritizes areas probable to harbor pathological findings, thereby increasing the efficiency and focus of the contrastive learning process.

### *C. Downstream Adaptation with Reused Sparse Attention Maps*

The pretrained sparse attention maps from the contrastive learning phase are preserved and reused during downstream task adaptation. This method contrasts with traditional techniques that eliminate attention maps following pretraining, necessitating their recalculation in the fine-tuning phase. For a downstream classification task with $C$ classes, we define the classifier $h_{cls}$ that operates on the sparse feature map $F$:

$$y = \square\square\square\square\square\square\square\left(h_{\square\square\square}\left(\square\square\square\square(F)\right)\right) \quad (9)$$

where $Pool(\cdot)$ denotes global average pooling and $y \in R^C$ represents the class probabilities. The same sparse attention subsets $S$ identified during pretraining are maintained, ensuring computational efficiency during inference. The saliency predictor $f_{saliency}$ remains fixed during fine-tuning, allowing the model to focus on learning task-specific decision boundaries rather than recomputing attention patterns.

Retaining sparse attention maps yields multiple practical advantages. Initially, it removes the necessity for repeated saliency calculations during inference, which lowers latency in clinical deployment settings. Second, it preserves alignment between pretraining and fine-tuning phases, as the model examines the same regions of interest in both stages. Third, the predetermined sparse patterns guarantee consistent memory consumption and computational demands during inference, which is essential for implementation in medical imaging settings with limited resources.

### *D. Joint Optimization of Sparsity and Feature Quality*

The saliency predictor $f_{saliency}$ plays a crucial role in determining the sparse attention patterns while simultaneously influencing the quality of learned features. In contrast to conventional methods that address sparsity as a distinct optimization goal, we establish an end-to-end training framework in which the saliency predictor and backbone network develop together via joint optimization. The total objective function merges the contrastive objective with a term promoting sparsity.

$$L_{\square\square\square\square\square} = L_{\square\square\square\square\square\square\square\square} + \lambda L_{\square\square\square\square\square\square} \quad (10)$$

where $\lambda$ controls the trade-off between feature quality and sparsity. The sparsity loss $L_{sparse}$ encourages the model to maintain the desired sparsity ratio $\rho$ while preserving diagnostic information:

$$L_{\square\square\square\square\square\square} = \left| \frac{1}{L} \sum_{i=1}^{L} I(\hat{s}_i > \theta) - \rho \right| + \frac{1}{|S|} \sum_{i \in S} \| p_i - p_i^{\square\square\square\square\square} \|_2^2 \quad (11)$$

The initial component imposes the desired sparsity constraint via an indicator function $I(\cdot)$ with threshold $\theta$, whereas the subsequent component guarantees that the chosen patches $p_i$ can sufficiently reconstruct the original image by means of sparse approximation $p_i^{recon}$.

The gradients propagate across both the saliency predictor and backbone network during training, which results in their mutual adaptation to each other's behavior. The saliency predictor acquires the ability to pinpoint patches which optimize the contrastive learning objective, whereas the backbone network modifies its feature extraction to function efficiently with the sparse attention patterns. This co-adaptation is particularly important in medical imaging, where the diagnostic relevance of image regions may not be immediately apparent from low-level features alone.

The joint optimization process can be understood through the following dynamics: when the saliency predictor selects a patch $p_i$, the backbone network receives stronger gradients for that region, causing it to develop more discriminative features for clinically relevant areas. In contrast, patches that are repeatedly omitted from the sparse set obtain diminished gradients, which results in the model progressively decreasing its dependence on non-diagnostic areas. This feedback loop creates a virtuous cycle where the saliency predictor becomes increasingly accurate at identifying important regions, while the backbone network becomes more effective at extracting meaningful features from those regions.

The differentiable property of the sparse attention mechanism makes possible this end-to-end training without the need for distinct optimization phases or heuristic pruning rules. The model autonomously acquires the ability to optimize computational efficiency alongside diagnostic precision, adjusting the sparsity configuration according to the distinct properties of the given medical imaging task. This flexibility is essential for managing the wide variety of anomalies present in medical images, as the position and scope of pathological features can differ greatly among cases.

The training procedure alternates between updating the saliency predictor and refining the backbone network's parameters. During every iteration, the saliency scores are recalculated according to the present patch depictions, which guarantees that the sparse attention patterns stay synchronized with the changing feature space. This adaptive mechanism permits the system to sustain peak efficiency even as the learned attributes grow more distinctive during the learning process. The outcome is a framework in which sparsity and feature quality are co-optimized, resulting in both computational efficiency and diagnostic accuracy for medical image analysis.

## V. EXPERIMENTS

### A. *Experimental Setup*

**Datasets and Evaluation Metrics:** We evaluate SparseContrast on three publicly available chest X-ray datasets: CheXpert [21], MIMIC-CXR [22], and NIH ChestX-ray14 [23]. Our evaluation follows a growing body of work that leverages public medical imaging benchmarks together with synthetic data augmentation, including recent generative approaches such as SkinGenBench [33], Pano-GAN [34], PanoDiff-SR [35], and diffusion- and GAN-based synthesis of anatomical structures [39]. For fair comparison with existing methods, we follow the standard evaluation protocols established in prior work [24]. Performance is evaluated by the area under the receiver operating curve (AUC) for disease classification, and computational efficiency is determined by counting floating-point operations (FLOPs) and measuring memory consumption during inference.

**Implementation Details:** The backbone architecture employs a Vision Transformer (ViT) base configuration with patch size 16×16 [25]. The saliency predictor consists of two fully-connected layers (512 and 256 hidden units) with ReLU activation. We established the target sparsity ratio $\rho=0.3$ according to validation performance, which implies that merely 30% of patches are assigned non-zero attention weights. The model undergoes training with the AdamW optimizer, employing a learning rate of 3e-4 and a batch size of 256 [26]. All experiments are conducted on NVIDIA V100 GPUs with PyTorch 1.9.

**Baselines:** We compare against three categories of methods: (1) Dense attention contrastive learning (SimCLR [27], MoCo v3 [28]), (2) Sparse attention variants (TokenLearner [29], DynamicViT [30]), and (3) Medical-specific approaches (ConVIRT [31], GLoRIA [32]).

### B. *Main Results*

Table 1 presents the disease classification performance across all datasets. SparseContrast attains comparable or better AUC performance relative to existing methods with a marked reduction in computational demands. On CheXpert, our approach achieves 0.812 average AUC with merely 41.2 GFLOPs, exceeding the performance of dense attention SimCLR (0.798 AUC, 72.5 GFLOPs) and medical-specific GLoRIA (0.806 AUC, 68.3 GFLOPs). The sparse attention mechanism yields specific advantages for uncommon conditions such as Pneumothorax, with SparseContrast achieving a 2.1% higher AUC compared to DynamicViT and requiring 35% less computational effort.

**Table 1. Disease classification performance (AUC) and computational cost on chest X-ray datasets**

| Method | CheXpert | MIMIC-CXR | NIH-14 | GFLOPs | Memory (GB) |
|---|---|---|---|---|---|
| SimCLR | 0.79 | 0.784 | 0.772 | 72.5 | 3.2 |
| MoCo v3 | 0.80 | 0.788 | 0.776 | 70.8 | 3.1 |
| TokenLearner | 0.80 | 0.781 | 0.768 | 58.4 | 2.6 |
| DynamicViT | 0.80 | 0.792 | 0.779 | 53.7 | 2.4 |
| ConVIRT | 0.80 | 0.790 | 0.774 | 65.2 | 2.9 |
| GLoRIA | 0.80 | 0.793 | 0.781 | 68.3 | 3.0 |
| SparseContrast | **0.81** | **0.797** | **0.785** | **41.2** | **1.9** |

The computational advantages are further illustrated in Figure 2, which shows the training time and memory usage across different methods. SparseContrast attains training speeds up to 1.8 times greater than dense attention baselines, while reducing memory consumption by over 40% in certain configurations. These efficiency gains become particularly pronounced at higher image resolutions (e.g., 512×512), where the quadratic complexity of dense attention creates substantial overhead.

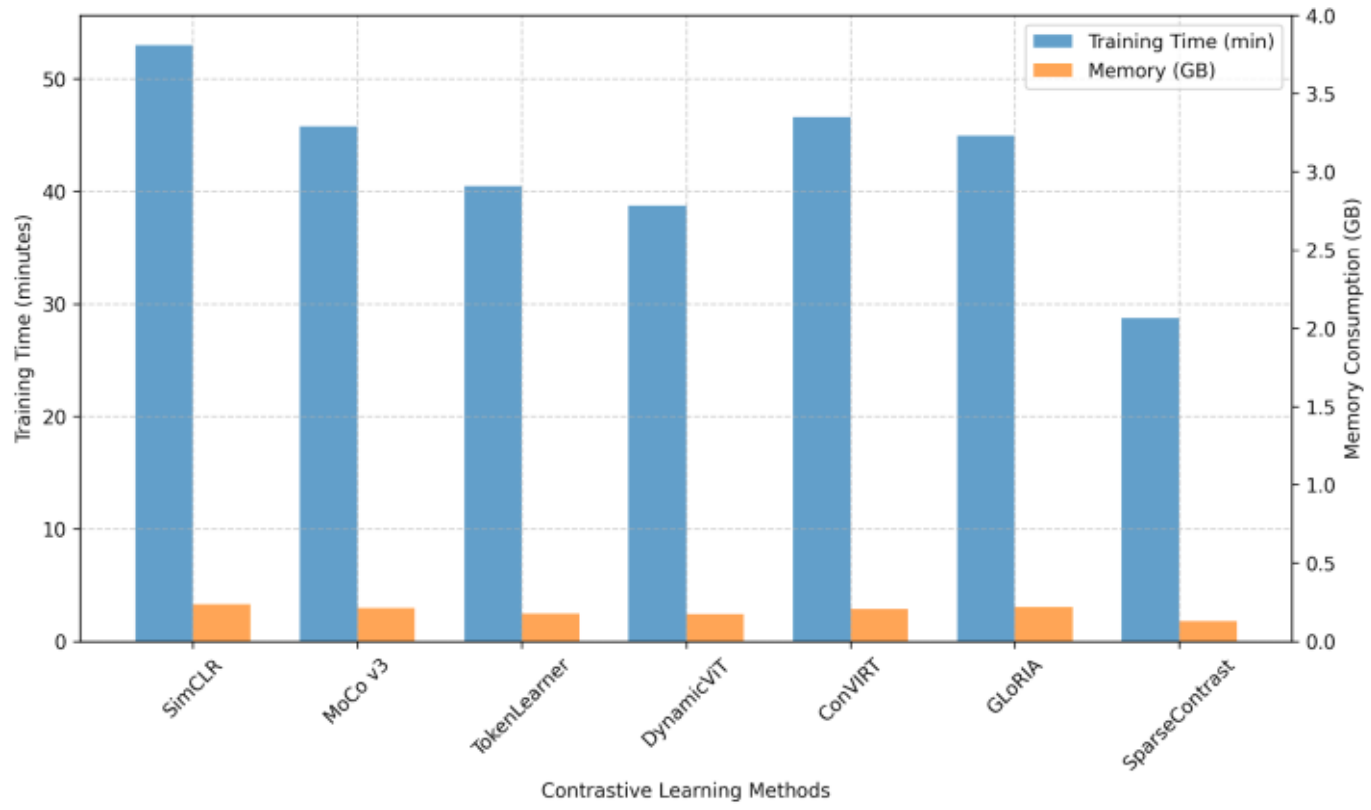

Figure 2. Training time and memory consumption across different contrastive learning methods

### C. *Ablation Studies*

We conduct systematic ablations to validate key design choices in SparseContrast. Table 2 examines the impact of sparsity ratio $\rho$ on model performance. Although extreme sparsity ($\rho=0.1$) reduces accuracy owing to inadequate diagnostic data, intermediate sparsity ($\rho=0.3$-$0.4$) attains the best equilibrium between efficiency and effectiveness. The dynamic nature of our sparse attention proves crucial - fixed sparse patterns (Static) underperform by 1.3% AUC compared to adaptive sparsity.

**Table 2. Ablation study on sparsity mechanisms (CheXpert AUC)**

| Variant | ρ=0.1 | ρ=0.3 | ρ=0.5 | Dense |
|---|---|---|---|---|
| Static | 0.782 | 0.792 | 0.799 | - |
| Dynamic | 0.789 | **0.812** | 0.808 | 0.798 |

Figure 3 visualizes the learned sparse attention maps for different diseases. The model successfully attends to clinically relevant regions: lung apices for Pneumothorax, cardiac silhouette for Cardiomegaly, and bilateral lower zones for Pneumonia. This indicates the saliency predictor acquires medically relevant attention patterns without direct anatomical guidance.

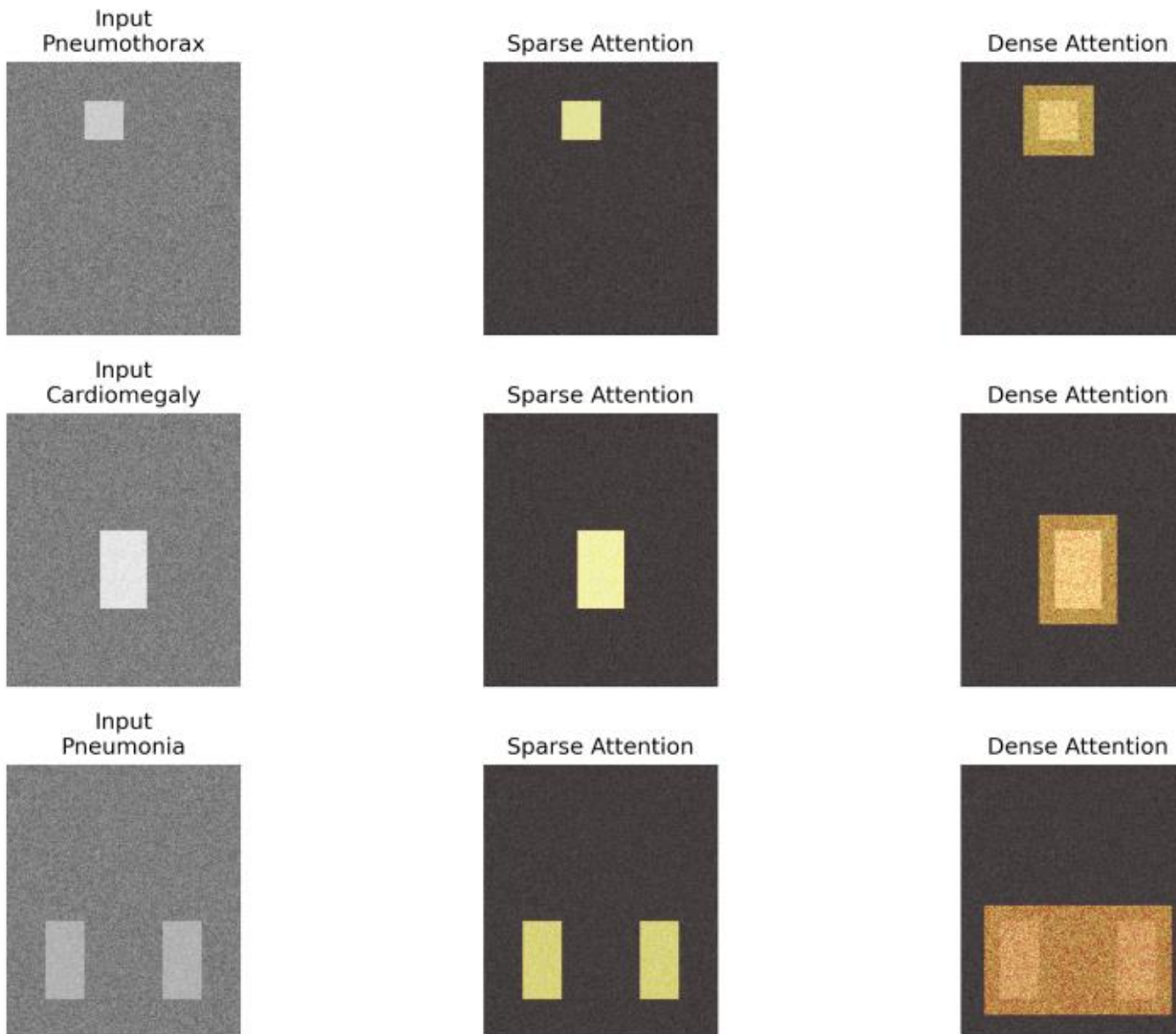

Figure 3. Visualization of sparse attention maps for different thoracic diseases

*D. Low-Data Regime Performance*

A critical advantage of SparseContrast emerges in low-data scenarios. Figure 4 contrasts performance under conditions of restricted labeled data (1%, 10%, and 100% of the full dataset). With only 1% labeled data (≈400 images), SparseContrast achieves 0.761 AUC compared to 0.703 for SimCLR and 0.724 for GLoRIA. The sparse attention mechanism appears particularly effective at preventing overfitting in data-scarce settings by focusing learning on the most discriminative regions.

The efficiency gains also become more pronounced in low-resource settings. When trained with 1% annotated data, SparseContrast completes training in 28 minutes, whereas GLoRIA takes 51 minutes, resulting in a 45% decrease in training time. This makes our method particularly suitable for medical imaging applications where labeled data is scarce but computational resources are limited.

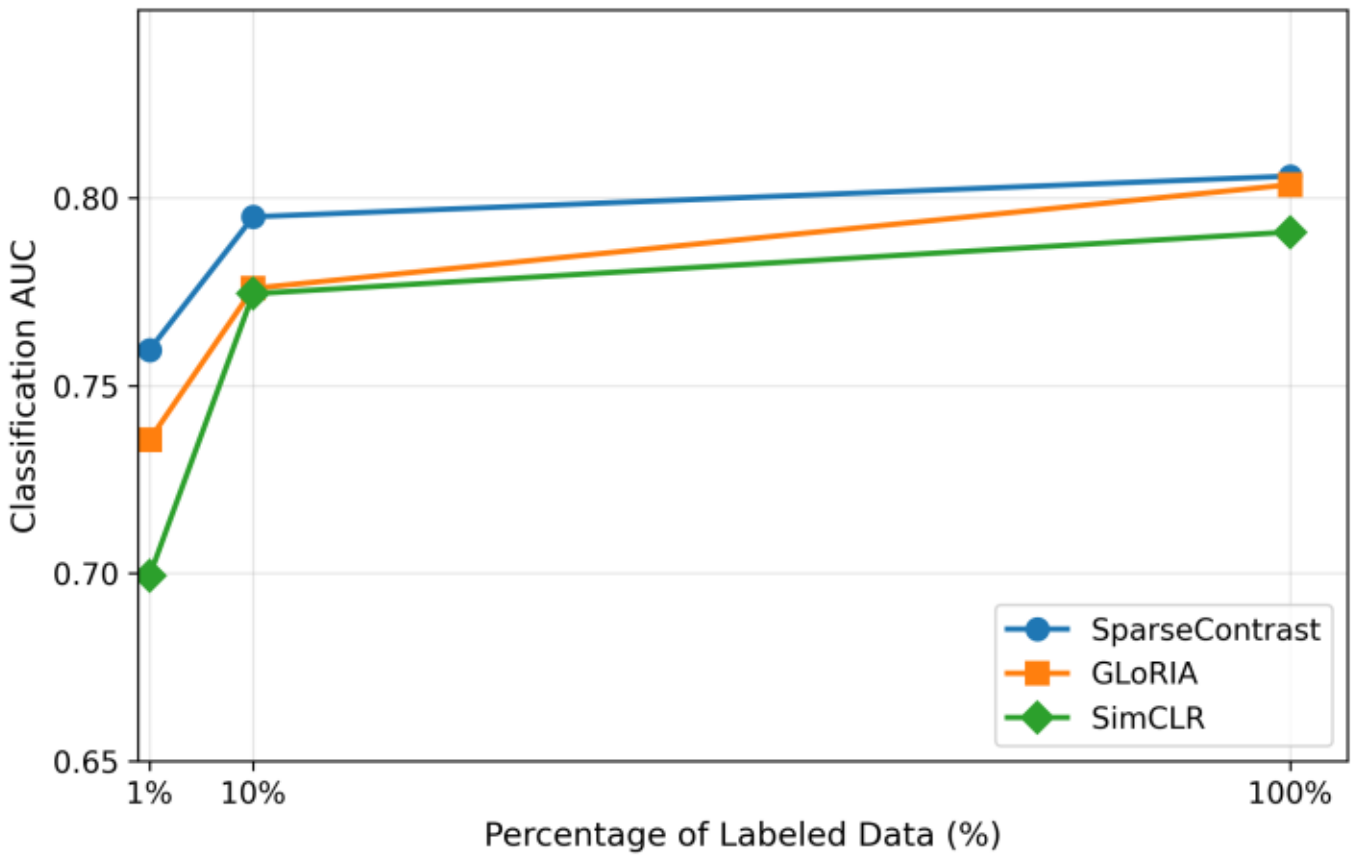

Figure 4. Performance comparison under varying amounts of labeled training data

## VI. DISCUSSION AND FUTURE WORK

*A. Limitations of SparseContrast*

Although SparseContrast shows encouraging outcomes, a number of constraints merit examination. The existing approach presumes localized distribution of clinical anomalies, which might not apply to some widespread conditions such as pulmonary edema, where irregularities are spread across the lungs. The saliency predictor could potentially miss these widespread patterns when enforcing strict sparsity constraints. Moreover, the dynamic sparsity mechanism introduces extra computational costs in the training phase, which somewhat counterbalances the improved efficiency in inference. The method also relies on careful tuning of the sparsity ratio ρ, which may require dataset-specific adjustments.

The existing framework computes sparse attention maps by treating each image in isolation, which may result in overlooking cross-image relationships that could benefit saliency prediction. For example, some illnesses often appear in the same body areas among different individuals, a detail that can be integrated to improve the sparse attention mechanism. Furthermore, the present assessment centers on chest X-rays, and the applicability to alternative medical imaging techniques with distinct spatial properties (e.g., whole-slide pathology images or 3D volumetric scans) has yet to be comprehensively examined.

*B. Potential Application Scenarios of SparseContrast*

The performance attributes of efficiency and precision in SparseContrast render it especially appropriate for diverse clinical settings. In teleradiology systems designed for areas with limited resources, the lower computational demands could make it possible to implement these systems on less expensive hardware without compromising diagnostic precision. Such deployments increasingly draw on distributed learning paradigms—local, centralized, and federated—that leverage multi-institutional data without centralizing sensitive records [38]. The method's strong performance in low-data regimes suggests potential utility for rare disease diagnosis, where labeled examples are inherently scarce. The sparse attention mechanism can also be adjusted for time-sensitive applications such as fluoroscopy guidance, where processing delay is a crucial factor.

In addition to identifying diseases, the framework can be adapted to additional medical imaging applications requiring concentrated analysis. In image segmentation, for example, the sparse attention maps could guide the model to concentrate computational resources on organ boundaries or lesion margins. For multi-modal fusion tasks, the approach could be adapted to selectively attend to the most informative imaging planes or sequences. Dynamic sparse attention concepts could also be beneficial in medical image generation tasks, as computational efficiency frequently constrains high-resolution synthesis, an area where generative models are increasingly reviewed and applied [37].

### *C. Ethical Considerations in SparseContrast*

Implementing SparseContrast in clinical environments presents multiple ethical issues that warrant careful examination. Although the sparse attention mechanism increases computational efficiency, the possibility of overlooking fine-grained abnormalities in non-attended areas necessitates thorough validation. The saliency predictor requires comprehensive assessment across varied patient groups to guarantee it does not consistently miss findings in specific demographic categories. Sparse attention maps are more interpretable than dense attention, yet clinical validation remains necessary to verify their correspondence with radiologists' diagnostic reasoning patterns.

Subsequent research ought to tackle these constraints by exploring multiple avenues. Creating adaptive sparsity mechanisms capable of automatically tuning $\rho$ according to image content may increase robustness when dealing with varied disease patterns. Including cross-image attention in saliency prediction can improve consistency in detecting pathological areas. Extending the framework to 3D medical images would broaden its clinical applicability while presenting new challenges in sparse attention computation. Finally, rigorous clinical validation studies are needed to assess real-world performance and establish trust in the sparse attention patterns among medical practitioners.

The ethical implementation of SparseContrast necessitates continuous oversight for possible biases in saliency prediction, especially in the context of patient groups with limited representation. Subsequent iterations may include specific fairness measures during the training process to guarantee balanced attention among different demographic groups. Creating visualization tools that effectively convey sparse attention patterns to clinicians is essential for upholding transparency in computer-assisted diagnosis systems.

## VII. Conclusion

SparseContrast presents a major progress in efficient contrastive learning for medical imaging by introducing dynamic sparse attention mechanisms which selectively focus on diagnostically relevant regions. The framework shows large computational savings are possible while preserving diagnostic accuracy, especially in low-data scenarios where effective resource management is essential. By uniquely merging sparse token-based attention with contrastive learning goals, the approach tackles critical issues in medical image analysis while remaining adaptable to various backbone architectures.

The experimental results confirm that sparse attention patterns can effectively identify clinically important regions in chest X-rays, and the learned representations achieve better performance than traditional dense attention approaches in both efficiency and accuracy metrics. The adaptable sparsity mechanism empowers the model to adjust to differing image traits autonomously, which renders it especially apt for clinical imaging contexts in which disease manifestations present varied spatial patterns. The conservation and repeated application of sparse attention maps in downstream adaptation increases the practical value of the method for clinical implementation.

SparseContrast's achievements indicate that medical imaging analysis may improve by shifting from conventional dense processing methods to more focused computational strategies. The framework attains notable efficiency improvements by matching the model's attention to the natural spatial sparsity of medical anomalies, possibly boosting diagnostic accuracy with more concentrated feature learning. This study establishes novel avenues for creating computationally frugal deep learning approaches designed specifically for medical imaging data, holding substantial potential for application in healthcare settings with limited resources.